%% file: paper.tex
\documentclass{article}

\makeatletter
\newcommand{\usepackagewithoutnatbib}[2][]{\@namedef{ver@natbib.sty}{9999/12/31}\let\setcitestyle\@gobble\usepackage[#1]{#2}\let\setcitestyle\undefined\expandafter\let\csname ver@natbib.sty\endcsname\@undefined}
\makeatother

\usepackagewithoutnatbib[preprint]{styles/icml2026}

\usepackage{hyperref}
\usepackage[center]{styles/VABcommands}
\usepackage{styles/VABenvironments}
\usepackage[authoryearcomp]{styles/VABcitations}

\usepackage{cleveref}
\usepackage{tikz}
\usepackage{subcaption}
\usepackage{csquotes}
\usepackage{booktabs}
\usepackage{standalone}
\usepackage{adjustbox}
\usepackage{array}
\usepackage{multirow}
\usepackage{nicefrac}
\MakeOuterQuote{"}

\usetikzlibrary{positioning}
\usetikzlibrary{arrows.meta}
\usetikzlibrary{calc}

\bibliography{references}

\hypersetup{
    colorlinks,
    linkcolor={red!50!black},
    citecolor={blue!50!black},
    urlcolor={blue!80!black}
}

\definecolor{greypointcolor}{rgb}{0.863,0.863,0.863}

\begin{document}

\twocolumn[
  
  \icmltitle{Bayesian Scattering: A Principled Baseline for Uncertainty on Image Data}
  \icmlsetsymbol{equal}{*}

  \begin{icmlauthorlist}
    \icmlauthor{Bernardo Fichera}{tue}
    \icmlauthor{Zarko Ivkovic}{eth}
    \icmlauthor{Kjell Jorner}{eth}
    \icmlauthor{Philipp Hennig}{tue}
    \icmlauthor{Viacheslav Borovitskiy}{uoe}

  \end{icmlauthorlist}

  \icmlaffiliation{tue}{University of Tübingen}
  \icmlaffiliation{eth}{ETH Zurich}
  \icmlaffiliation{uoe}{University of Edinburgh}

  \icmlcorrespondingauthor{B. Fichera}{bernardo.fichera@gmail.com}
  \icmlcorrespondingauthor{V. Borovitskiy}{viacheslav.borovitskiy@gmail.com}


  \vskip 0.3in
]

\printAffiliationsAndNotice{Code available at: \href{https://github.com/nash169/bayesian-scattering}{github.com/nash169/bayesian-scattering}}

\begin{abstract}
Uncertainty quantification for image data is dominated by complex deep learning methods, yet the field lacks an interpretable, mathematically grounded baseline.
We propose \emph{Bayesian scattering} to fill this gap, serving as a first-step baseline akin to the role of Bayesian linear regression for tabular data. 
Our method couples the wavelet scattering transform---a deep, non-learned feature extractor---with a simple probabilistic head.
Because scattering features are derived from geometric principles rather than learned, they avoid overfitting the training distribution.
This helps provide sensible uncertainty estimates even under significant distribution shifts.
We validate this on diverse tasks, including medical imaging under institution shift, wealth mapping under country-to-country shift, and Bayesian optimization of molecular properties.
Our results suggest that Bayesian scattering is a solid baseline for complex uncertainty quantification methods.
\end{abstract}

\begin{figure*}[t]
\centering

\resizebox{\linewidth}{!}{
  \input{figures/intro/figure}
}
\caption{Bayesian scattering maps an input image $\m{X}$ to a probabilistic prediction $p(y \given \Phi(\m{X}))$ via the scattering features $\Phi(\m{X})$.}
\label{fig:intro}
\end{figure*}
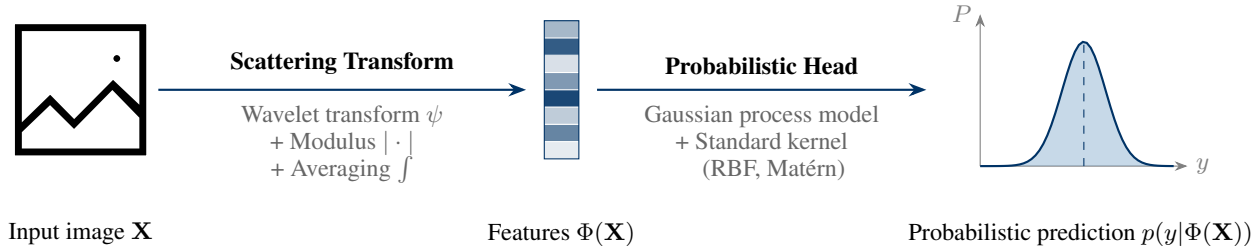

\section{Introduction}
Uncertainty estimates are crucial for automated decision-making, safety-critical applications, and more \cite{abdar2021, gawlikowski2023}.
Yet, deep learning methods, while highly accurate, often fail to deliver useful uncertainty estimates, especially under distribution shift---when test-time inputs differ substantially from the training ones~\cite{koh2021, gustafsson2023}.\footnotemark[2]

Furthermore, although recent breakthroughs in machine learning have largely been enabled by massive datasets and ever-larger models, many problems in domains of growing importance---such as scientific and medical applications---lack the luxury of abundant data.
While transfer learning and foundation models offer a potential remedy, they introduce new complexities regarding safety, robustness, and uncertainty calibration \cite{awais2025, bommasani2021}.
We argue that this increasing complexity necessitates interpretable baselines to anchor empirical evaluation and clarify the true benefits of more complex approaches.
Establishing such a baseline for image data is the main objective of this paper.

\footnotetext[2]{In this paper, \emph{distribution shift} refers to \emph{covariate shift}: when the input distribution changes while the conditional remains fixed.\vspace{-\baselineskip}}

We propose the \emph{scattering transform}~\cite{bruna2013} as the foundation for such a baseline.
Scattering transforms produce deep, hierarchical feature maps that are provably translation-invariant and stable to small deformations and additive noise~\cite{mallat2010}.
These properties explicitly encode the geometric inductive biases of natural images---biases that convolutional neural networks and vision transformers must otherwise acquire through massive pretraining~\cite{dosovitskiy2020}.
This grounding makes the scattering transform a principled choice, while its nature as a fixed, non-learned feature extractor ensures it cannot overfit the training distribution, thereby offering inherent robustness to distribution shift.
Its effectiveness as a powerful feature extractor has been demonstrated in computer vision---particularly in data-scarce regimes~\cite{oyallon2018, gauthier2022}---and in scientific modeling tasks~\cite{hirn2017, eickenberg2017}.

Despite this record, the potential of scattering transforms for uncertainty quantification has been largely overlooked.
To fill this gap, we propose \emph{Bayesian scattering}, which combines scattering features with a simple probabilistic head such as a Gaussian process \cite{rasmussen2006}, as illustrated in~\Cref{fig:intro}.
We describe how this method arises from the geometry of the image space and discuss its main properties.
By coupling a principled, geometry-motivated feature extractor---immune to overfitting by design---with a simple probabilistic head, we create an interpretable and robust reference model.
Crucially, it is not intended to claim state-of-the-art performance, but rather to constitute a principled baseline for evaluating more complex approaches.
Such a baseline is particularly important in uncertainty quantification, where scalar metrics may be insufficient to guarantee safety or downstream utility, necessitating a reference model that is grounded by design rather than merely optimized for a specific score.

In our experiments, we focus on regression, as it allows exact Gaussian process inference in small data settings, thereby minimizing confounding factors in method evaluation.
To provide practical guidelines, we first conduct a preliminary study in a simple, stylized setting (\texttt{QM2D} dataset of~\textcite{hirn2017}), analyzing hyperparameter sensitivity and feature behavior.
We then demonstrate the method's efficacy on diverse tasks with realistic distribution shifts, including two medical imaging settings with institution shifts \cite{gustafsson2023}, as well as wealth mapping \cite{yeh2020, koh2021} under a country-to-country shift.
Finally, we validate the utility of the baseline in downstream decision-making via Bayesian optimization of molecular properties on the \texttt{QM2D} and \texttt{QM9} datasets~\cite{ruddigkeit2012, ramakrishnan2014}.

\section{Related Work}

\paragraph{Uncertainty for images.}
The standard approach to uncertainty quantification for images involves modifying or post-processing deep neural networks.
Prominent methods include deep ensembles~\cite{lakshminarayanan2017}, which average predictions over multiple independently trained models; Monte Carlo dropout~\cite{gal2016}, which approximates Bayesian inference via dropout at test time; and scalable Laplace approximations~\cite{daxberger2021}, which estimate the posterior curvature around a MAP solution.
While effective, these methods rely on learned feature representations that can be opaque and prone to overconfidence under distribution shift~\cite{ovadia2019,gustafsson2023}.
Hybrid approaches like deep kernel learning~\cite{wilson2016} or other combinations of neural networks and Gaussian processes~\cite{bradshaw2017} attempt to mitigate this by placing probabilistic heads on top of deep networks.
However, they still inherit the data-hunger and biases of the underlying learned features.

\vspace{-\baselineskip}

\paragraph{Scattering transforms and applications.}
The scattering transform~\cite{bruna2013} has been proposed as a robust feature extractor for small-data regimes in image classification~\cite{oyallon2018} and scientific modeling~\cite{hirn2017, eickenberg2017}.
While several engineering studies have technically combined scattering features with Gaussian processes~\cite{ferkous2021, sudirman2022, lone2023, ojha2024}, they do not constitute a precedent for the baseline we propose.
Crucially, most of these works~\cite{ferkous2021, sudirman2022, lone2023} utilize the Gaussian process solely for kernel regression (predicting the mean) and omit uncertainty quantification entirely.
The only exception~\cite{ojha2024} considers uncertainty but is restricted to time series rather than image data and concentrates on the specific application of impact localization.
To our knowledge, no prior work has proposed or explicitly evaluated Bayesian scattering as a robust baseline for uncertainty quantification on image data.

\section{Scattering}
\label{sec:scattering}

We now introduce \emph{wavelet scattering transforms}, or simply \emph{scattering transforms}, which form the backbone of the principled baseline we propose.
We begin by reviewing the fundamental properties desirable for any feature representation of images.
These properties, which capture the geometric inductive biases of natural images, constitute the defining characteristics of scattering transforms.
We then detail the construction of the transform from cascades of \emph{linear wavelet transforms}, pointwise \emph{nonlinearities}, and \emph{pooling} by (local) averaging.
Finally, we discuss the integration of rotation invariance in scattering transforms and the dimensionality of the resulting scattering feature maps.

\subsection{Defining Properties of Scattering Transforms}
\label{sec:scattering:defining_properties}

To motivate scattering transforms, we first discuss the general properties that any ``good'' feature map for images should satisfy.
For simplicity, let us neglect the discrete nature of images for a while and treat them as functions \mbox{$f: \R^2 \to \R$.
Let $\Phi: L^2(\R^2) \to \mathcal{H}$ denote our feature map.}

\begin{figure}[t]
  \centering
  
  \newcommand{\figimg}[2]{%
    \includegraphics[height=#1, width=1.65cm, keepaspectratio]{#2}%
  }

  \tikzset{
    frame/.style={
        draw=black!20, line width=0.5pt, inner sep=0pt,
        minimum width=1.65cm, minimum height=1.75cm,
        anchor=center, align=center
    },
    arrow/.style={
        -Latex, draw=black!70, thick, line width=1.0pt, 
        shorten <=4pt, shorten >=4pt
    },
    node distance=0.5cm
  }

  \begin{subfigure}[b]{0.48\linewidth}
    \centering
    \begin{tikzpicture}
      \node[frame] (A) {\figimg{1.68cm}{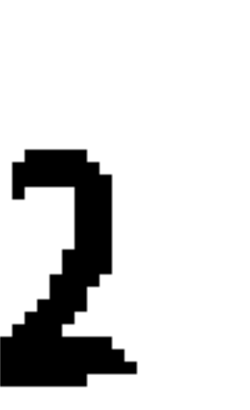}};
      \node[frame, right=of A] (B) {\figimg{1.68cm}{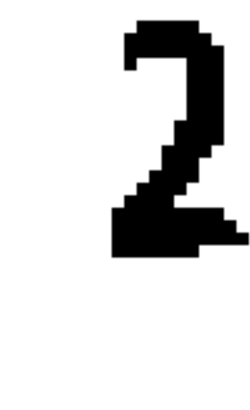}};
      \draw[arrow] (A) -- (B);
    \end{tikzpicture}
    \caption{Translation}
    \label{fig:transforms:translation}
  \end{subfigure}\hfill
  \begin{subfigure}[b]{0.48\linewidth}
    \centering
    \begin{tikzpicture}
      \node[frame] (A) {\figimg{1.10cm}{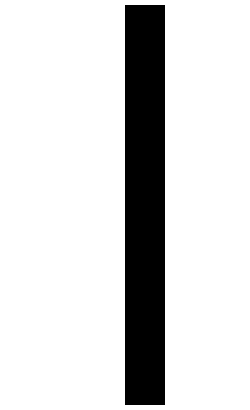}};
      \node[frame, right=of A] (B) {\figimg{1.10cm}{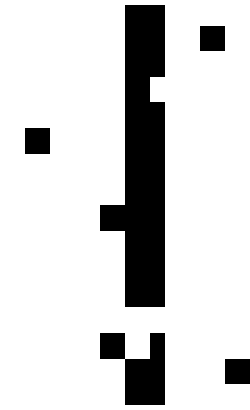}};        
      \draw[arrow] (A) -- (B);
    \end{tikzpicture}
    \caption{Additive noise}
    \label{fig:transforms:additive_noise}
  \end{subfigure}

  \par\bigskip

  \begin{subfigure}[b]{1.0\linewidth}
    \centering
    \begin{tikzpicture}
      \node[frame] (A) {\figimg{1.10cm}{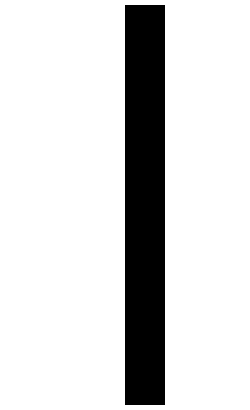}};
      \node[frame, right=of A] (B) {\figimg{1.10cm}{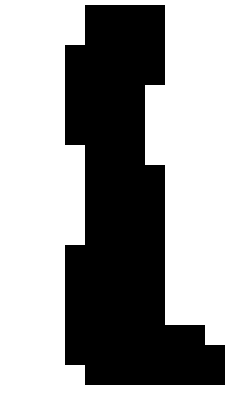}};
      \node[frame, right=of B] (C) {\figimg{1.10cm}{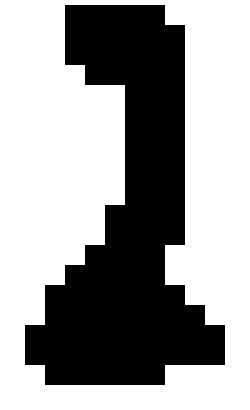}};
      \node[frame, right=of C] (D) {\figimg{1.10cm}{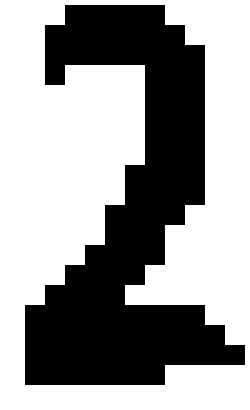}};
      
      \draw[arrow] (A) -- (B) (B) -- (C) (C) -- (D);
    \end{tikzpicture}
    \caption{Deformation}
    \label{fig:transforms:deformation}
  \end{subfigure}
  
  \caption{Different types of transformations that feature maps should be invariant or stable to. Note: while single steps in (\subref{fig:transforms:deformation}) are small deformations, their composition changes the image content.}
  \label{fig:transforms}
\end{figure}
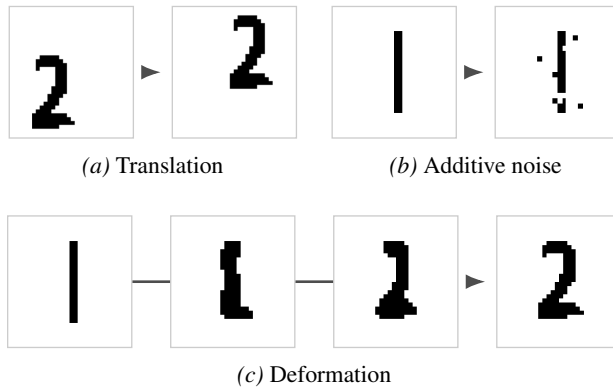

First, since shifting an image does not change its semantic content, the feature map $\Phi$ should satisfy
\begin{enumerate}
    \item[1.] \textbf{Translation-invariance:} $\Phi(T_{\v{c}} f) = \Phi(f)$ where we have $f \in L^2(\R^2)$, $\v{c} \in \R^2$, and $T_{\v{c}} f(\v{x}) = f(\v{x} - \v{c})$.
\end{enumerate}
Translation-invariance (\Cref{fig:transforms:translation}) is important, being a key inductive bias of convolutional neural networks.\footnote{Although the convolutions themselves are \emph{equivariant} \cite{cohen2016}, the presence of subsampling---e.g., as a result of pooling---in traditional architectures leads to (local) invariance. As discussed later in the paper, scattering transforms can also be made to exhibit  local, rather than global, invariance.}

Can we further simplify the feature space by demanding invariance to a broader class of transformations?
Deformations---formally, $C^2$-diffeomorphisms $\tau \!: \! \R^2 \!\!\!\to\! \R^2$ which act on $f: \R^2 \to \R$ via $(D_{\tau} f)(\mathbf{x}) = f(\tau(\mathbf{x}))$---form a much richer group than translations.
However, invariance to all such deformations would eliminate essential information, as illustrated in~\Cref{fig:transforms:deformation}.
A more appropriate requirement is \emph{stability to small deformations}: small, smooth input variations should yield only small changes in the feature representation.
This property is formalized as
\begin{enumerate}
    \item[2.] \textbf{Stability to small deformations:}
    for all $f \in L^2(\R^2)$ and for all $C^2$-diffeomorphisms $\tau$ with $\norm{\tau - \mathrm{id}} \leq 1$, it is true that
    $\norm{\Phi(D_{\tau} f) - \Phi(f)} \leq C \norm{f} \norm{\tau - \mathrm{id}}$.
    Here, $\mathrm{id}(\v{x}) = \v{x}$ is the identity; $C > 0$ is a constant.
\end{enumerate}

In addition to deformation stability, a feature map should naturally be stable to additive noise (see~\Cref{fig:transforms:additive_noise}):
\begin{enumerate}
    \item[3.] \textbf{Stability to additive noise:}
    for all $f, \varepsilon \in L^2(\R^2)$ and some $C > 0$, we have
    $\norm{\Phi(f + \varepsilon) - \Phi(f)} \leq C \norm{\varepsilon}$.
\end{enumerate}

Finally, the feature map must be \emph{rich} enough for downstream learning: it should not eliminate crucial information or collapse semantically meaningful distinctions.
Ideally, one seeks features that are as informative as possible, potentially invertible up to invariances such as translation.

Constructing feature maps that are simultaneously invariant, stable, and expressive turns out to be a highly nontrivial task.
Naïve approaches often succeed in only one or two of these aspects, but not all of them simultaneously.
\emph{Wavelet scattering transforms} are specifically designed to address this challenge.
For a discussion of typical failure modes in naïve approaches and further motivation, see~\Cref{appdx:scattering}.

\subsection{Explicit Construction}
\label{sec:scattering:explicit_construction}

This section provides an explicit construction of scattering transforms, focusing on the concrete steps and intuition most relevant for practical applications.
For clarity and brevity, we deliberately omit a number of important technical assumptions, as well as subtleties concerning convergence and normalization.
All of these are rigorously addressed in~\textcite{mallat2012, bruna2013}.

\paragraph{Wavelets}
We start by fixing a family of "nice" complex-valued \emph{wavelets} $\psi_{j, r} : \R^2 \to \C$, where $j \in \Z$ encodes the scale and $r$ indexes a finite group $R \subset \mathrm{O}(2)$ of rotations and reflections.
These are simple waveforms, localized both in frequency and spatial domains, chosen so that their Fourier transforms $\widehat{\psi_{j,r}}$ cover the whole frequency domain.
Then, the \emph{wavelet transform} maps $f$ to $\cbr{W[j, r] f}_{j \in \Z, r \in R}$, where $W[j, r] f = f * \psi_{j, r}$.
Each $\psi_{j, r}$ is a convolutional filter, but here it is analytically defined rather than learned.

\paragraph{Translation-invariant scattering transform}
To build rich representations, \textcite{mallat2012} proposed to combine wavelet transforms with nonlinearities---specifically, with the complex modulus~$\abs{\cdot}$, leading to $\mathrm{U}[j, r] f = \abs{W[j, r] f}$, and then cascade these $\mathrm{U}[j, r]$, building deep representations.
For a \emph{path} $p = ((j_1, r_1), \ldots, (j_m, r_m))$ this leads to the definition of the \emph{scattering propagator} $\mathrm{U}[p]$:
\[
\label{eqn:propagator}
\mathrm{U}[p] f = \mathrm{U}[j_m, r_m] \ldots \mathrm{U}[j_1, r_1] f
\]
Each $\mathrm{U}[p]$ is stable to additive noise and small deformations \cite{mallat2012}.
Furthermore, the collection $\{\mathrm{U}[p] f\}_{p}$ is maximally expressive, allowing recovery of $f$.
However, $\mathrm{U}[p]$ is not translation-invariant---rather, it is translation-equivariant \cite{mallat2012}.
Furthermore, it is overly redundant and impractical to work with, being an infinite collection of functions.
The \emph{translation-invariant scattering transform} maps $f$ to $\cbr{\mathrm{S}[p] f}_{p}$ where $\mathrm{S}[p] f: L^2(\R^2) \to \R$ is built from $\mathrm{U}[p]$ by averaging---in the context of neural networks, a form of pooling---of the scattering propagator:
\[
\mathrm{S}[p] f = \int_{\R^2} \mathrm{U}[p] f(\v{x}) \d \v{x} \in \R
\]
In practice, one chooses a finite set of paths $P$---we discuss the choice in~\Cref{sec:scattering:dimensionality,sec:bs:instantiating}---and only computes $\mathrm{S}_P f = \cbr{\mathrm{S}[p]}_{p \in P} f \in \R^{\abs{P}}$.
It is guaranteed to satisfy all the properties from~\Cref{sec:scattering:defining_properties} \cite{bruna2013}.

\paragraph{Windowed scattering transform}
Sometimes it may be desirable to only retain invariance to \emph{small} translations, say only to translations by $\norm{\v{c}} \lll 2^J$, $J \in \N$.
This is achieved by the \emph{windowed scattering transform} $\mathrm{S}^w$.
It replaces global averaging with a localizing convolution,
\[
\mathrm{S}^w[p] f = \mathrm{U}^w[p] f * \phi_{2^J},
\quad
\phi_{2^J}(\mathbf{x}) = 2^{-dJ} \phi(2^{-J}\mathbf{x}),
\]
where $\phi$ is a smooth, rapidly-decaying bump function (e.g., a Gaussian). 
Importantly, unlike $\mathrm{S}[p]$, which outputs a scalar, $\mathrm{S}^w[p]$ outputs a function which in practice has to be discretized.
To preserve the properties 2 and 3 from~\Cref{sec:scattering:defining_properties}, the propagator $\mathrm{U}^w$ should be redefined by restricting onto paths with $j \leq J$ and allowing the averaging $f * \phi_{2^J}$ to be used alongside any $\mathrm{U}[j, r] f$ in~\Cref{eqn:propagator}.

\subsection{Rotation Invariance}
\label{sec:rotation_invariance}

Whether rotation invariance is desirable for a feature map depends on the task: for example, it is not always appropriate in digit recognition (e.g., a $6$ rotated by $180^\circ$ resembles a $9$), but important for predicting molecular properties from approximate electronic density images of arbitrary orientation.
With a suitable wavelet design, scattering transforms are \emph{equiva\textbf{}riant} to rotations~\cite{hirn2017, eickenberg2017}.
In this case, a rotation-invariant propagator $\mathrm{S}^{\text{rot-inv}}[(j_1, \ldots, j_m)]$ can be obtained by averaging:
\[ \label{eqn:rotation_invariance}
\frac{1}{|R|^m} \sum_{r_m \in R} \cdots \sum_{r_1 \in R}
    \mathrm{S}[(j_1, r_1), \ldots, (j_m, r_m)]f.
\]

\subsection{Feature Map Dimensionality}

\label{sec:scattering:dimensionality}

The expressiveness of scattering features depends on the number and structure of retained paths.
\textcite{mallat2012, waldspurger2017} show that $|\mathrm{S}[p]f|$ decay exponentially with path length $m$.
It thus suffices to restrict to short paths, typically $m \leq 2$ or $3$.
Moreover, only paths with $j_1 < \cdots < j_m$ and $1 \leq j_i \leq J$ are really needed, where $J$ is a hyperparameter with $2^J \leq N$ for discrete images of size $N \times N$.
There are $|R|^m \binom{J}{m}$ paths of length $m$, so truncating at $m \leq M$ gives $\sum_{m=1}^M |R|^m \binom{J}{m}$ features.
For windowed scattering, this total is multiplied by the output resolution, typically $N^2 / 2^{2J}$ for $N \x N$ images.
With rotation invariance~(\Cref{eqn:rotation_invariance}), the $\abs{R}^m$ factors drop out.

\section{Bayesian Scattering}
\label{sec:bs}

We now formalize \emph{Bayesian scattering}, a framework that pairs the fixed, principled features of the scattering transform with a simple probabilistic head.
The core premise is that while simple probabilistic models---such as Gaussian processes---excel on tabular data, they are ill-suited for the high dimensionality and complex geometry of raw images.
The scattering transform bridges this gap: it reduces dimensionality and filters irrelevant variation, projecting images into a structured feature space where the Euclidean metric aligns with semantic content better.
This mirrors the simplification of learned representations, but with a key advantage: because scattering features are not optimized on the training set, they remain independent of the training distribution.
This decoupling avoids overfitting and fosters robust uncertainty estimation even under distribution shift.

\subsection{General Framework}

Consider a dataset $\c{D} = \{(\m{X}_i, y_i)\}_{i=1}^n$, where inputs $\m{X}_i$ lie in pixel space $\c{I} = \R^{N \x N}$ and outputs $y_i \in \c{Y}$ represent targets (e.g., $\c{Y}=\R$ for regression, $\c{Y}=\{0, \dots, C-1\}$ for classification).\footnote{For multi-channel images (e.g., RGB), scattering is applied to each channel independently; resulting features are concatenated.}
Let $\Phi : \c{I} \to \R^M$ be the fixed scattering feature map.
We define a probabilistic model over the target $y$ given the transformed input via some latent function $f$:
\[
y \mid \m{X}, f \sim p(y \mid f(\Phi(\m{X})))
,
\]
where $p(y \mid \cdot)$ is an appropriate likelihood.
By placing a prior $p(f)$ over functions on the feature space $\R^M$, we effectively induce a prior $f^{\Phi}(\m{X}) \coloneqq f(\Phi(\m{X}))$ on the original pixel space $\c{I}$.
Given the dataset $\c{D}$, the posterior distribution over the latent function is $p(f \mid \c{D})$, while the predictive for a new test point $\m{X}^*$ is obtained by marginalization:
\[
p(y^* \mid \m{X}^*, \c{D}) = \int p(y^* \mid f(\Phi(\m{X}^*))) \, p(f \mid \c{D}) \, df.
\]

This predictive distribution encapsulates both the deterministic prediction and uncertainty.
For regression~(${\c{Y}=\R}$), the expectation $\E[y^*]$ corresponds to the prediction, while the standard deviation $\sqrt{\Var[y^*]}$ quantifies uncertainty---intuitively, the estimated magnitude of the error.
For classification, these correspond to the mode (predicted class) and the entropy of the distribution, respectively.

\subsection{Instantiating the Framework}
\label{sec:bs:instantiating}

To instantiate this framework, we adhere to standard scattering conventions~\cite{bruna2013, hirn2017, eickenberg2017}.
First, we restrict to paths of length $m \leq 2$: deeper paths are typically unnecessary due to the exponential decay of feature energy and computationally prohibitive due to growth in feature dimensionality.
Second, following the implementation in \texttt{Kymatio}~\cite{andreux2020}, when using the windowed scattering transform, we discretize the output with a stride of $2^J$, yielding a feature map of spatial resolution $N/2^J \times N/2^J$.

Within this setup---and assuming a wavelet family is fixed by the implementation---Bayesian scattering involves the following three modeling choices:
\begin{enumerate}
    \item[\textbf{1.}] \textbf{Scattering invariances:} Use the fully translation-invariant transform $\mathrm{S}$ or its windowed version $\mathrm{S}^w$; whether to impose rotation-invariance using $\mathrm{S}^{\text{rot-inv}}$.
    \item[\textbf{2.}] \textbf{Scattering hyperparameters:} Number of scales $J$ ($2 \leq 2^J \leq N$), size of the discrete rotation group $L=|R|$---or similar in case of non-standard wavelets.
    \item[\textbf{3.}] \textbf{Probabilistic head:} This includes the prior, which can come with hyperparameters to be fit from data, and the inference method for obtaining the posterior.
\end{enumerate}

The choice between $\mathrm{S}$, $\mathrm{S}^w$, and $\mathrm{S}^{\text{rot-inv}}$ is usually guided by the invariances inherent to the problem at hand.
Moreover, scattering hyperparameters can typically be treated in the same way as the hyperparameters of the probabilistic head.  

\paragraph{Probabilistic head}
To maintain the transparency of the baseline, we strongly recommend keeping the probabilistic head simple.
Specifically, we advocate for Gaussian process (GP) regression or classification with standard covariance functions, such as the Matérn, RBF, or linear kernels~\cite{rasmussen2006}.
Note that the latter choice recovers Bayesian linear or logistic regression.
A standard advantage of Gaussian processes is the ability to perform model selection (e.g., kernel choice, or even choice of $J$) by simply maximizing the \emph{log marginal likelihood} on the training data.
While this theoretically streamlines the workflow by removing the need for a validation set, we caution that it can lead to overfitting for the relatively high-dimensional feature spaces typical of scattering transforms.

\subsection{Inference and Computation}
\label{sec:bs:inference}

Because the scattering feature map $\Phi$ is fixed, feature extraction is strictly decoupled from the probabilistic modeling.
This separation allows to move the data from the complex, high-dimensional pixel space $\mathcal{I}$ into a more structured, moderate-dimensional vector space $\R^M$, effectively reducing the task to a standard probabilistic problem on tabular data.
Consequently, the choice of inference method is dictated solely by computational constraints (dataset size $N$) and the target task (likelihood $p(y \mid \cdot)$).

\paragraph{Feature extraction}
Computationally, extracting scattering features is equivalent to a forward pass of a fixed convolutional network involving wavelets and smooth bump functions, as well as the modulus non-linearities.
This process is highly efficient and amenable to standard GPU acceleration, supported by libraries such as \texttt{Kymatio}~\cite{andreux2020}, which handles both 2D images and their 3D counterparts.
The resulting feature map dimension was discussed in~\Cref{sec:scattering:dimensionality}.
In practice, the dimensionality of the scattering feature map usually falls between $10^2$--$10^4$. 

\paragraph{Exact inference}
For regression tasks on moderate-sized datasets ($N$ of order $10^3$), we recommend exact Gaussian process regression.
In this regime, the learning problem reduces to standard Gaussian process regression on transformed inputs $\v{x}_i = \Phi(\m{X}_i) \in \R^M$.
This approach provides a ``gold-standard'' baseline devoid of inference-level approximation errors, where the posterior mean and variance are computed via Cholesky decomposition with $\mathcal{O}(N^3)$ complexity~\cite{rasmussen2006}.
Crucially, the log marginal likelihood can be computed exactly with the same complexity, further streamlining model selection.

\paragraph{Variational inference}
To scale to larger $N$, or to handle the non-Gaussian likelihoods required for classification, one can employ approximate inference.
Specifically, the stochastic variational Gaussian process (SVGP) framework of \textcite{hensman2013} introduces a set of \emph{inducing inputs} $Z \subset \R^M$ and corresponding \emph{inducing variables} $\v{u} = f(Z)$.
The joint posterior is approximated by a variational distribution with the structure $q(f, \v{u}) = p(f \mid \v{u}) q(\v{u})$.
The parameters of $q(\v{u})$ are learned by maximizing the evidence lower bound (ELBO), which factorizes over the data:
\[
\sum_{i=1}^N
\E_{f \sim q(f)}
\log p(y_i \mid f(\v{x}_i))
-
\mathrm{KL}[q(\v{u}) \,||\, p(\v{u})]
.
\]
This factorization enables mini-batch training similar to standard deep learning and reduces complexity to $\mathcal{O}(\abs{Z}^3)$ where $\abs{Z} \ll N$.
Such flexibility enables Bayesian scattering to be used for large-scale or classification tasks.

\begin{figure*}[t]
    \centering
    \sffamily\footnotesize
    
    \newcommand{\pair}[2]{%
        \includegraphics[width=1.85cm, height=1.85cm, keepaspectratio]{#1}\hspace{1.5pt}%
        \includegraphics[width=1.85cm, height=1.85cm, keepaspectratio]{#2}%
    }
    
    \newcommand{\rothead}[1]{%
        \raisebox{\dimexpr 0.925cm - 0.5\height \relax}{\rotatebox{90}{\textbf{#1}}}%
    }

    \setlength{\tabcolsep}{0pt}
    \begin{tabular*}{\textwidth}{c @{\hspace{6pt}} c @{\extracolsep{\fill}} c c c c}
        
        & \textbf{HistologyNucleiPixels} & \textbf{SkinLesionPixels} & \textbf{AssetWealth} & & \textbf{Molecule Representations} \\
        
        \cmidrule(lr){2-2} \cmidrule(lr){3-3} \cmidrule(lr){4-4} \cmidrule(lr){6-6}
        
        \rothead{Train} & 
        \pair{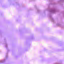}{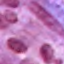} &
        \pair{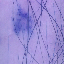}{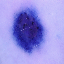} &
        \pair{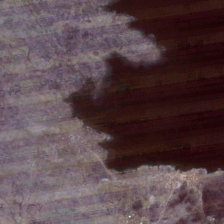}{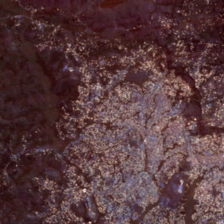} & &
        \pair{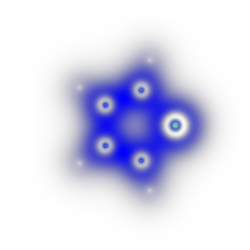}{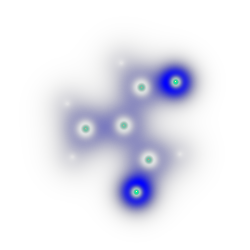} \\
        
        \addlinespace[6pt]
        
        \rothead{Test} & 
        \pair{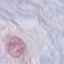}{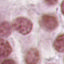} &
        \pair{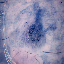}{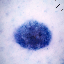} &
        \pair{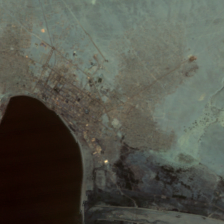}{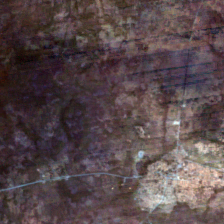} & &
        \pair{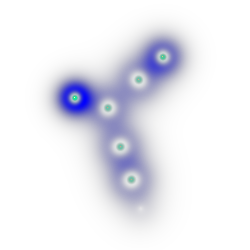}{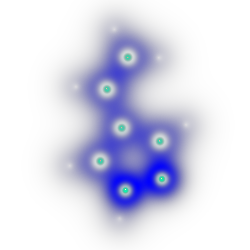} \\
        
        \addlinespace[6pt]
        
        & \multicolumn{3}{c}{(a) Distribution Shift Benchmarks} & & (b) QM2D (No Shift) \\
    \end{tabular*}

    \caption{Datasets used in~\Cref{sec:experiments} with the exception of QM9 whose 3D image representations are difficult to visualize.}
    \label{fig:experiments_inputs}
    \vspace*{-\baselineskip}
\end{figure*}

\section{Experiments}
\label{sec:experiments}

We evaluate Bayesian scattering across three distinct regimes to validate its utility as a robust baseline.
First, we conduct a sensitivity analysis in a controlled setting to establish best practices and understand hyperparameter trade-offs.
Second, we assess performance under real-world distribution shifts using three diverse regression benchmarks: histology, skin lesion analysis, and satellite-based wealth prediction.
Finally, we investigate the utility of the method for downstream decision-making via Bayesian optimization of molecular properties.

\subsection{Sensitivity Analysis and Best Practices}
\label{sec:experiments:sensitivity}

Before addressing distribution shifts, we conduct a preliminary study to establish practical guidelines for Bayesian scattering.
We focus on \texttt{QM2D}, a dataset of 4,357 planar molecules derived from the standard \texttt{QM7} dataset~\cite{blum2009, rupp2012}.
Following~\textcite{hirn2017}, we represent each molecule as a 2D image with three channels: valence electronic density, core electronic density, and "Dirac"---in~\Cref{fig:experiments_inputs}~(b) we visualize these as RGB images.
While this task does not involve distribution shift, it serves as a controlled test for understanding hyperparameter sensitivity and computational trade-offs.
Detailed results are in~\Cref{appdx:experiments}; here we summarize the key takeaways that inform our subsequent experiments.

\paragraph{Scattering configuration}
We study the impact of the number of scales $J$ and the number of angular orientations $L = \abs{R}$ (the size of the rotation group).
Here, we consistently found that $L=8$ provides the optimal balance between performance and computation; increasing $L$ beyond this point yielded no performance gains.
For the scaling parameter, $J \approx \log_2 N - 1$ (where $N$ is as in $N \x N$ resolution image) proved optimal across resolutions.
This suggests that, in practice, retaining some local translation sensitivity may be beneficial compared to the full global averaging ($J = \log_2 N$) even for fully translation-invariant~tasks.

\paragraph{Feature dimensionality reduction}
Contrary to common practice in scattering literature~\cite{bruna2013}, we found that applying principal component analysis (PCA) to scattering features---even when keeping 100\% of variance---often degraded both calibration and accuracy.
We hypothesize that the rotation of the feature space disrupts its geometric structure, resulting in a representation less aligned with the priors induced by standard kernels.
Furthermore, truncating the basis to retain only partial variance can be detrimental for uncertainty quantification under distribution shift, as it may discard features that are latent in the training distribution but vital for generalization.
We thus recommend using raw scattering features.

\paragraph{Kernel choice and ARD}
The parameter-free linear kernel frequently exhibited numerical instability in our experiments; we therefore recommend standard RBF or Matérn kernels.
While automatic relevance determination (ARD) kernels~\cite{mackay1992,rasmussen2006} can improve the flexibility of the model by learning per-feature lengthscales, we found them prone to overfitting as feature dimensionality grows.
For this reason, we recommend ARD kernels primarily for lower-dimensional feature maps (e.g., when rotation-invariance or full translation-invariance is imposed) and advise exercising caution when applying them to higher-dimensional representations.

\paragraph{Optimization and efficiency}
For exact Gaussian process inference, we found that optimizing the log marginal likelihood for 500 iterations using Adam~\cite{kingma2014} provides a robust trade-off between convergence and cost.
To ensure numerical stability, we recommend initializing lengthscales to the mean pairwise distance of the training features.
In terms of wall clock time, feature extraction for QM2D (at $J=7, L=8$) took approximately 876 seconds with rotation invariance and 1236 seconds without; we attribute the latter's increased runtime to implementation specifics.
Compared to the 790 seconds required for a small 3.7M parameter CNN (ConvNeXt-Atto) on the same hardware, these results demonstrate that Bayesian scattering is competitive with small deep learning models.

\subsection{Image Regression}
\label{sec:experiments:regression}

\begin{table*}[t]
	\centering
	\caption{Performance comparison on image regression tasks under real-world distribution shifts. We report the mean and standard deviation across 5 synchronized random splits. The best result is \underline{underlined}; results not statistically significantly different are \textbf{bolded}.}
	\label{tab:main_results}
	\small
	\setlength{\tabcolsep}{0pt} 
	\begin{tabular*}{\linewidth}{@{\extracolsep{\fill}} llccccccc}
		\toprule
		\multirow{2}{*}[-0.5ex]{\textbf{Data}} & \multirow{2}{*}[-0.5ex]{\textbf{Metric}} & \multirow{2}{*}[-0.5ex]{\textbf{Trivial}} & \multirow{2}{*}[-0.5ex]{\textbf{Ensemble}} & \multirow{2}{*}[-0.5ex]{\textbf{Scattering}} & \multicolumn{2}{c}{\textbf{ConvNeXt}} & \multicolumn{2}{c}{\textbf{DINOv2}} \\
		\cmidrule(lr){6-7} \cmidrule(lr){8-9}
		& & & \smash{\scriptsize (5x ConvNeXt-Atto)} & & \textbf{Atto} & \textbf{Base} & \textbf{Small} & \textbf{Base} \\
		\midrule

		\multirow{5}{*}{\textbf{SLP}}
        & RMSE            & 815 $\pm$ 34 & 626 $\pm$ 29 & $\mathbf{543 \pm 28}$ & \underline{$\mathbf{541 \pm 40}$} & $\mathbf{558 \pm 21}$ & 783 $\pm$ 39 & 730 $\pm$ 34 \\
        & NLL             & 1.52 $\pm$ 0.06 & 12.49 $\pm$ 3.5 & $\mathbf{1.35 \pm 0.19}$ & 1.62 $\pm$ 0.27 & \underline{$\mathbf{1.24 \pm 0.11}$} & 1.87 $\pm$ 0.15 & 1.80 $\pm$ 0.23 \\
        & QCE             & 0.03 $\pm$ 0.02 & 0.49 $\pm$ 0.05 & 0.14 $\pm$ 0.02 & 0.14 $\pm$ 0.02 & \underline{$\mathbf{0.08 \pm 0.02}$} & 0.14 $\pm$ 0.01 & 0.14 $\pm$ 0.02 \\
        & PI-$\mu$        & 3.92 $\pm$ 0.00 & 0.65 $\pm$ 0.05 & \underline{$\mathbf{1.62 \pm 0.08}$} & $\mathbf{1.64 \pm 0.10}$ & 2.24 $\pm$ 0.18 & 2.65 $\pm$ 0.09 & 2.59 $\pm$ 0.15 \\
        & PI-$\sigma$ $\uparrow$    & 0.00 $\pm$ 0.00 & 0.43 $\pm$ 0.05 & 0.64 $\pm$ 0.05 & 0.50 $\pm$ 0.06 & 0.30 $\pm$ 0.02 & 1.14 $\pm$ 0.13 & 1.26 $\pm$ 0.09 \\

        \midrule

		\multirow{2}{*}{\textbf{HP}}
        & RMSE            & 635 $\pm$ 19 & 475 $\pm$ 13 & \underline{$\mathbf{392 \pm 46}$} & 490 $\pm$ 50 & 513 $\pm$ 26 & $\mathbf{447 \pm 48}$ & $\mathbf{447 \pm 24}$ \\
        & NLL             & 1.33 $\pm$ 0.04 & 11.00 $\pm$ 3.21 & \underline{$\mathbf{0.82 \pm 0.07}$} & $\mathbf{0.88 \pm 0.09}$ & 0.96 $\pm$ 0.09 & $\mathbf{0.89 \pm 0.14}$ & $\mathbf{0.84 \pm 0.10}$ \\

		\midrule

		\multirow{2}{*}{\textbf{AW}}
        & RMSE            & 0.90 $\pm$ 0.02 & 0.63 $\pm$ 0.04 & 0.64 $\pm$ 0.03 & $\mathbf{0.57 \pm 0.04}$ & $\mathbf{0.56 \pm 0.03}$ & \underline{$\mathbf{0.53 \pm 0.03}$} & $\mathbf{0.53 \pm 0.02}$ \\
        & NLL             & 1.54 $\pm$ 0.03 & 7.77 $\pm$ 0.88 & 1.06 $\pm$ 0.05 & $\mathbf{0.98 \pm 0.07}$ & $\mathbf{0.98 \pm 0.07}$ & \underline{$\mathbf{0.92 \pm 0.06}$} & $\mathbf{0.93 \pm 0.05}$ \\

		\bottomrule
	\end{tabular*}
    \vspace*{-\baselineskip}
\end{table*}

We now evaluate Bayesian scattering on a set of regression tasks with real-world distribution shifts.

\paragraph{Datasets}
We focus on three datasets from the benchmarks of~\textcite{koh2021,gustafsson2023}
whose sample inputs we visualize in~\Cref{fig:experiments_inputs}~(a).
These~are

\begin{itemize}
    \item \textbf{SkinLesionPixels}~(SLP)~\cite{gustafsson2023}:
    Medical imaging task. The goal is to predict the count of lesion pixels in a $64 \times 64$ dermatoscopic image.
    It is based on the \texttt{HAM10000} dataset by~\textcite{tschandl2018} which consists of different sub-datasets, the training set (6,592 images) was collected in Austria while the test (2,259) was collected in Australia, with different equipment and skin type distribution.

    \item \textbf{HistologyNucleiPixels}~(HP)~\cite{gustafsson2023}:
    Medical imaging task. The goal is to predict the count of nuclei pixels in $64 \times 64$ histology crops.
    The shift is \emph{institutional}: the training set (10,808 images) is derived from \texttt{ConNSeP}~\cite{graham2019} data, while the test (2,267) comes from \texttt{TNBC}~\cite{naylor2018}, created by a different institution with inherent variations in staining and sample preparation.

    \item \textbf{AssetWealth}~(AW)~\cite{yeh2020, koh2021}:
    Wealth mapping task using 8-channel satellite images to predict an asset wealth index for locations across 23 African countries.
    We resize the original $224 \times 224$ inputs to $256 \times 256$ for compatibility with standard architectures.
    The setup defines a challenging distribution shift by splitting data based on country: the training set (11,797 images) and test set (3,963 images) are drawn from disjoint lists of nations.
    This introduces significant variation, as both the terrain and the visual markers of wealth differ across regions.
\end{itemize}

\paragraph{Metrics}
We report the negative log likelihood (NLL) to assess the joint quality of prediction and uncertainty.
Furthermore, to disentangle these aspects, we also report root mean squared error (RMSE) for predictive accuracy, quantile coverage error (QCE) to evaluate calibration, as well as mean and standard deviation of prediction interval width (PI-$\mu$ and PI-$\sigma$) to evaluate sharpness and dispersion---see~\textcite{tran2020} for definitions.

\begin{table}[b]
\centering
\caption{Ablation study on \texttt{SkinLesionPixels} (mean over 5 runs). We compare rotation-invariant Bayesian scattering (BS-RI) across scales $J$ and kernels (RBF $\pm$ ARD) against deep ensembles trained on scattering features and raw \texttt{ConvNeXt} features.}
\label{tab:ablation_kernels}
\small
\setlength{\tabcolsep}{0pt} 

\begin{tabular*}{\linewidth}{@{\extracolsep{\fill}} lcccccc}
\toprule
\multirow{2}{*}[-0.5ex]{\textbf{Metric}} & \multicolumn{2}{c}{\textbf{BS-RI ($J=3$)}} & \multicolumn{2}{c}{\textbf{BS-RI ($J=5$)}} & \multicolumn{2}{c}{\textbf{Ensemble}} \\
\cmidrule(lr){2-3} \cmidrule(lr){4-5} \cmidrule(lr){6-7}
 & \textbf{RBF} & \textbf{RBF-A} & \textbf{RBF} & \textbf{RBF-A} & \textbf{Scatt.} & \textbf{NN} \\
\midrule

Train Loss & 0.65 & \textbf{0.60} & 0.74 & 0.68 & - & - \\
RMSE       & 543 & 542 & 584 & 583 & \textbf{512} & 524 \\
NLL        & \textbf{1.34} & 1.37 & 1.78 & 2.0 & 18.6 & 37.8 \\

\bottomrule
\end{tabular*}
\end{table}

\paragraph{Setup}
To simulate critical data-scarce regimes, we randomly subsample training and test sets to $N_{\text{train}}=1000$ and $N_{\text{test}}=250$.
This protocol enables robust evaluation across five randomized, synchronized splits.
We benchmark Bayesian scattering ($L=8$) against exact GP regression on features from state-of-the-art pre-trained models: \texttt{ConvNeXt} (Atto 3.7M, Base 88M) and \texttt{DINOv2} (Small 22M, Base 88M).
For the 8-channel \texttt{AssetWealth} data, we adapt these RGB models by processing channels independently and concatenating the outputs.
To contextualize these against a standard learned baseline, we also evaluate a deep ensemble of five \texttt{ConvNeXt}-Atto models trained end-to-end. We selected the Atto variant as its smaller parameter count is the most appropriate for our data-scarce regime.
GP hyperparameters are optimized via log marginal likelihood; we report statistics for the best-performing configuration (kernel type, ARD, scale $J$, rotation-invariance) for each method.

\paragraph{Results}
The main results are summarized in \Cref{tab:main_results}.
Remarkably, all methods employing fixed features with a GP head drastically outperform the end-to-end deep ensemble across all metrics.
On \texttt{HP}, Bayesian scattering achieves top-tier performance, matching or exceeding the learned features in both accuracy (RMSE) and likelihood (NLL) while offering superior calibration.
On \texttt{SLP}, it remains highly competitive, outperforming \texttt{DINOv2} and rivaling \texttt{ConvNeXt}.
Crucially, detailed analysis (see \Cref{appdx:experiments}) reveals that scattering models here are "selectively uncertain"---producing prediction intervals that vary significantly in width depending on input difficulty---whereas neural baselines tend to output more static intervals.
Finally, on the \texttt{AW} task, while scattering lags behind deep models in raw predictive error, it maintains robust calibration (QCE) comparable to the competition.
This confirms that even when fixed geometric features limit predictive accuracy, the Bayesian scattering framework can still obtain reliable uncertainty.

\subsubsection{More Data and Approximate Inference}

To demonstrate scalability, we evaluate performance on the full \texttt{SLP} training set ($N=6,592$) using SVGP~\cite{hensman2013} inference with $\abs{Z}=1024$ inducing points and mini-batch training.

Results (see \Cref{appdx:experiments}, \Cref{tab:more_data}) confirm that the trends observed in the data-scarce regime persist at a larger scale.
While predictive accuracy naturally improves with more data, the primary benefit is significantly improved calibration (lower QCE and NLL).
This suggests that Bayesian scattering can effectively leverage denser data coverage to refine its uncertainty estimates.

\subsubsection{Model Choice and Deep Ensemble Heads}

Finally, we perform an ablation study to isolate the impact of the probabilistic head and verify whether training marginal likelihood reliably predicts generalization under distribution shift.
We benchmark the Gaussian process head against a deep ensemble of 5 MLPs (one hidden layer, 128 units) trained on the fixed scattering features.

Key results are summarized in \Cref{tab:ablation_kernels}.
Hyperparameter sensitivity proves domain-dependent: while \texttt{HP} is virtually invariant to configuration changes (see \Cref{appdx:experiments}), \texttt{SLP} is sensitive to both scale $J$ and the use of ARD.
Regarding the probabilistic head, deep ensembles on scattering features yield competitive mean predictions (RMSE) on \texttt{SLP} but suffer from extreme overconfidence (high NLL) under distribution shift.
They perform similarly to or worse than GPs on other datasets (\Cref{appdx:experiments}).
Critically, we find that training marginal likelihood is a poor proxy for model selection here.
For instance, ARD models on \texttt{SLP} achieve lower training loss but significantly higher test NLL compared to non-ARD kernels, highlighting the risk of overfitting when optimizing flexible lengthscales in high-dimensional geometric feature spaces.

\subsection{Bayesian Optimization}

We conclude our evaluation by applying Bayesian scattering to a downstream decision-making task: Bayesian optimization (BO) of molecular properties.
This setting is directly relevant to scientific discovery, where uncertainty estimates can guide sequential selection of candidates to optimize a target property efficiently~\cite{frazier2018}.

\paragraph{Planar molecules (QM2D)}
We begin with the \texttt{QM2D} dataset used in our sensitivity analysis (\Cref{sec:experiments:sensitivity}), where the task is to optimize quantum chemical energy (QE) of planar molecules.
This serves as a natural extension of our regression experiments, validating whether performance observed there translates to decision-making applications.

\paragraph{3D molecules (QM9)}
Since most realistic molecules are not planar, we extend our evaluation to the 3D domain using the full QM9 dataset~\cite{ramakrishnan2014}.
Here, we utilize the solid harmonic wavelet scattering transform~\cite{eickenberg2017, eickenberg2018}, which generalizes the principles of stability and invariance to volumetric data.
We optimize isotropic polarizability (IP), electronic spatial extent (ESE), and internal energy at 0K (U0).

\paragraph{Setup}
We employ a GP regression head.
While we tested various configurations (including ESE, U0, and different kernels; see~\Cref{appdx:experiments}), in the main text we restrict to Matérn-$\nicefrac{5}{2}$ kernel without ARD, which yielded the most consistent performance, and IP property.
We benchmark against \texttt{Uni-Mol}~\cite{zhou2023}, a pre-trained SE(3)-equivariant transformer designed specifically for molecular representation learning and not based on image-like representations of molecules.
Starting with 50 random molecules, we perform 50 iterations of sequential selection using expected improvement acquisition optimized over a pool of 1,000 random candidates.
Model hyperparameters are re-optimized at every step to adapt to the growing dataset.
We report \emph{simple regret}, defined as the difference between the true optimum and the best value found so far.

\begin{figure}[t]
	\centering
	\subfloat[][QE on \texttt{QM2D}, 2D setting]{\includegraphics[height=3.9cm]{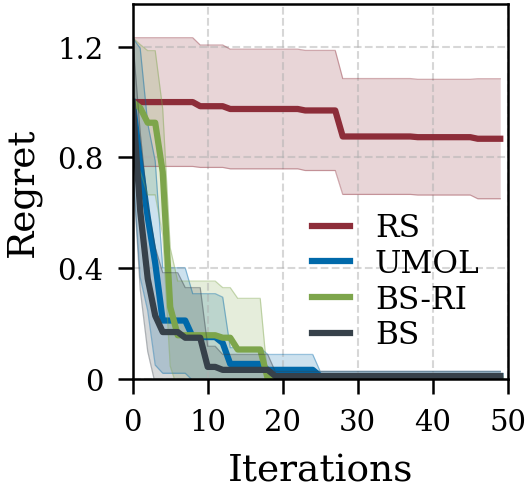}\label{fig:bo_comparison:2d}}
    \hfill
    \subfloat[][IP on \texttt{QM9}, 3D setting]{\includegraphics[height=3.9cm]{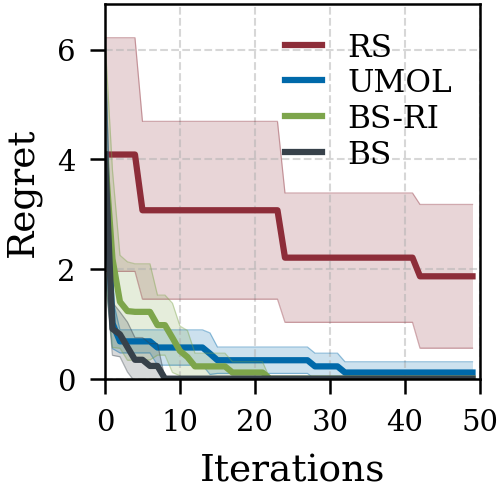}\label{fig:bo_comparison:3d}}
    \caption{
    Regret of BO of molecular properties.
    Curves show the mean over 5 runs; shaded regions show $0.95$ confidence intervals.
    In the legend, \texttt{RS} stands for random search, \texttt{UMOL} for \texttt{Uni-Mol}, \texttt{BS} for Bayesian scattering and \texttt{RI} for rotation-invariant.
    }
    \label{fig:bo_comparison}
\end{figure}

\paragraph{Results}
\Cref{fig:bo_comparison} shows the optimization trajectories.
On 2D molecules (\texttt{QM2D}), Bayesian scattering (BS) demonstrates remarkable efficiency, converging to near-zero regret in roughly 10 iterations while \texttt{Uni-Mol} requires somewhat more (\Cref{fig:bo_comparison:2d}).
This performance confirms that for standard tasks involving planar geometry, a principled baseline can be surprisingly competitive.
In the 3D setting (\texttt{QM9}), solid harmonic scattering proves equally effective (\Cref{fig:bo_comparison:3d}).
Contrary to the assumption that voxel-based representations might struggle with complex quantum properties, Bayesian scattering performs on par with---and arguably converges faster than---the domain-specific \texttt{Uni-Mol} baseline.
This demonstrates that the geometric priors encoded in the scattering transform provide a robust foundation for decision-making even in complex settings.

\section{Conclusion}
We present Bayesian scattering as a principled baseline for uncertainty quantification for images.
By combining the geometric inductive biases of the scattering transform with simple probabilistic heads, it avoids the overfitting typical of learned features, fostering robustness to distribution shift.
Empirically, we show that this baseline is competitive with deep learning methods in data-scarce regression and effective for downstream decision-making.
We hope this work provides a rigorous foundation to ground the evaluation of future robust machine learning systems.

\printbibliography

\newpage

\onecolumn
\appendix
\crefalias{section}{appendix}

\section{Scattering Transforms}
\label{appdx:scattering}

\subsection{Failure Modes in Satisfying the Properties from~\Cref{sec:scattering:defining_properties}}

Constructing feature maps that are simultaneously invariant, stable, and expressive is a nontrivial challenge. Naïve choices often fail to balance these:

\begin{itemize}
    \item The identity map $\Phi(f) = f$ is maximally expressive, but lacks translation-invariance.
    \item Canonical invariants---such as aligning an image based on its maximum---achieve translation-invariance, but become highly unstable to small deformations or additive noise.
    \item The absolute value of the Fourier transform is also translation-invariant, and stable to additive noise, but---as shown in~\textcite{mallat2012}---is \emph{not} stable to small deformations.
    \item Convolutions can be made stable to small deformations and additive noise, but they are translation-equivariant ($\Phi(T_{\v{c}} f) = T_{\v{c}} \Phi(f)$) rather than invariant ($\Phi(T_{\v{c}} f) = \Phi(f)$).
    \item Globally pooling equivariant features such as convolutions, e.g. by averaging, produces invariance but often at the expense of expressivity, potentially discarding crucial information.
\end{itemize}

These examples illustrate the tension between invariance, stability, and richness in feature design. The wavelet scattering transform resolves this trade-off by interleaving convolution, non-linearity, and averaging in a principled cascade, yielding features that provably satisfy all three desiderata.

\section{Additional Experimental Details}
\label{appdx:experiments}

In this appendix, we report the additional results for the different scenarios evaluated in the paper, as well as provide more details of experimental setups.

\subsection{Sensitivity Analysis and Best Practices}

As discussed in~\Cref{sec:experiments:sensitivity}, an extensive empirical evaluation was conducted to study the sensitivity of Bayesian scattering to various design choices and come up with best practices.
Taking inspiration from \textcite{hirn2017}, we used a subset of the QM7 dataset of small molecules consisting of only planar molecules for these tests.
The resulting dataset comprises a total of 4,357 molecules and is termed \texttt{QM2D} below.  
Following the methodology of \textcite{hirn2017}, for each molecule we generated a 2D image representation with three channels corresponding to the electronic density of the valence electrons, core electrons, and to the \emph{dirac} channel which places atomic charge values at the locations of the respective nuclei.

The files \texttt{QM2D\_1---5.pdf} in \url{https://github.com/nash169/bayesian-scattering/tree/main/results} report a range of experiments designed to study the sensitivity of model performance to various design choices.
Specifically, we analyze the impact of: (1) training set size and number of training iterations; (2) scattering transform hyperparameters, i.e., the number of scales \( J \) and the number of angular orientations \( L \); (3) number of input channels; (4) image resolution; (5) application of PCA to the scattering features.

Complementing the analysis in~\Cref{sec:experiments:sensitivity}, we observe that increasing the number of input channels consistently improves performance, as expected given the additional information they provide.
Notably, however, the valence channel emerges as the most influential input.
This aligns with chemical intuition, suggesting that valence interactions are the primary determinants of the relevant molecular properties.
We omit these details from the main text because channel importance---while a valid sanity check---is highly specific to this dataset and offers limited generalizable insight.

\subsection{Image Regression}

Detailed tabulated results for the image regression experiments are provided in the file \texttt{dist-shift-exact-gp.pdf} in \url{https://github.com/nash169/bayesian-scattering/tree/main/results}.
In this section, we provide a dataset-specific analysis of the performance trends and hyperparameter sensitivity.
For all image regression tasks, we employed a consistent evaluation protocol: images were resized to $64 \times 64$ pixels (allowing a maximum scattering scale $J=6$), and models were trained on 1000 points and evaluated on 250 points using 5 (potentially intersecting) splits randomly subsampled from the dataset and subordinate to the dataset-predefined split (train is subsampled from the train part only, test is subsampled from the test part only, where the original train and test parts exhibit a shift), with splits synchronized across all methods.
We evaluated scattering transforms with $L=8$ angles, considering both standard and rotation-invariant features coupled with RBF and Matérn kernels (with and without ARD).

\subsubsection{\texttt{SkinLesionPixels} Dataset}

For this dataset, we evaluated scattering scales $J \in \{3, 4, 5\}$.
For each scale we evaluated rotation covariant and invariant scattering features.

\paragraph{Performance Analysis.}
On the skin lesion task, Bayesian scattering remains competitive but presents a distinct trade-off profile compared to the neural features. Interestingly, DinoV2 performs worse on this task by virtually all accounts.
\begin{itemize}
    \item \textbf{RMSE:} Scattering achieves an RMSE comparable to both ConvNeXt Base and ConvNeXt Atto but is considerably better than DinoV2.
    \item \textbf{NLL:} Scattering achieves an NLL that is worse but comparable to ConvNeXt Base, while outperforming ConvNeXt Atto. The NLL of DinoV2 is considerably worse.
    \item \textbf{Calibration (QCE):} In terms of calibration error, scattering is strictly worse than ConvNeXt Base (statistically significant) and on par with ConvNeXt Atto. DinoV2 is considerably worse than scattering in QCE.
    \item \textbf{Prediction Intervals:} Despite the somewhat higher QCE, scattering exhibits desirable uncertainty characteristics for decision-making. Its prediction intervals are much narrower than those of ConvNeXt Base, yet they exhibit significantly higher variance from point to point. This variability suggests the model is selectively uncertain, whereas ConvNeXt Base tends to output broad, uniform intervals regardless of input difficulty.
\end{itemize}

\paragraph{Model Selection and Sensitivity.}
In contrast to the histology dataset, performance here is highly sensitive to hyperparameters.
Rotation-invariant scattering with $J=3$ ($L=8$) performs significantly better than all other scattering configurations.
Regarding the probabilistic head, non-ARD kernels consistently outperform ARD versions, while the choice between RBF and Matérn kernels has negligible impact.
This dataset also provides a clear example where training objectives fail to predict test performance: ConvNeXt Base yielded one of the highest (worst) training losses ($0.948 \pm 0.053$) yet achieved the best test NLL, confirming that model selection based on training loss is unreliable here.
Finally, although ensembles as probabilistic heads achieve top performance in mean prediction here, they suffer from massive overconfidence.

\subsubsection{\texttt{HistologyNucleiPixels} Dataset}

For this dataset, we evaluated scattering scales $J \in \{3, 4, 5\}$.
For each scale we evaluated rotation covariant and invariant scattering features.

\paragraph{Performance Analysis.}
Bayesian scattering demonstrates top-tier performance on the histology dataset, matching or exceeding the results of neural features across both deterministic and probabilistic metrics.
\begin{itemize}
    \item \textbf{RMSE:} Scattering achieves an RMSE on par with both DinoV2 variants and strictly outperforms ConvNeXt.
    \item \textbf{NLL:} In terms of negative log-likelihood, scattering is generally superior or comparable to all neural network models.
    \item \textbf{Calibration (QCE):} Scattering shows strong calibration; its QCE is comparable to the best-performing neural model (ConvNeXt Atto) and is significantly better than the rest of neural features.
    \item \textbf{Prediction Intervals:} Regarding the geometry of uncertainty, scattering produces prediction intervals that are narrower (more informative) than competitors for the respective QCE levels. However, on the other hand, the standard deviation of these interval widths is relatively small across the test set.
\end{itemize}

\paragraph{Model Selection and Sensitivity.}
Performance on this dataset is highly robust to hyperparameter variations.
Metrics (excluding PI statistics) are largely insensitive to the scattering scale $J$, the use of rotation invariance, the choice of kernel (RBF vs. Matérn), or the use of ARD.
Minor differences appear only in PI statistics, where we observe a slight preference for $J=5$ without rotation invariance using an RBF kernel (regardless of ARD).
Training marginal likelihood values were very comparable across models, making them an ineffective metric for model selection.
Finally, using ensembles as probabilistic heads proved detrimental, yielding worse performance across all metrics compared to alternative using Gaussian processes.

\subsubsection{\texttt{AssetWealth} Dataset}

For scattering, we resized the images to $256 \times 256$, and evaluated scattering scales $J \in \{5, 6, 7\}$.
For each scale we evaluated rotation covariant and invariant scattering features.

\paragraph{Performance Analysis.}
In this domain, Bayesian scattering generally lags behind learned features.
\begin{itemize}
    \item \textbf{RMSE:} Scattering yields significantly higher error than both ConvNeXt and DinoV2 variants.
    \item \textbf{NLL:} In terms of negative log-likelihood, scattering performs comparably to ConvNeXt but is significantly outperformed by DinoV2.
    \item \textbf{Calibration (QCE):} Despite lower predictive accuracy, the calibration of Bayesian scattering remains robust, achieving QCE scores comparable to both ConvNeXt and DinoV2.
    \item \textbf{Prediction Intervals:} Scattering tends to produce wider, less informative prediction intervals compared to the neural baselines, although the dispersion (standard deviation) of interval widths is slightly more favorable.
\end{itemize}

\paragraph{Model Selection and Sensitivity.}
Hyperparameter choices play a critical role here.
We find that ARD is highly detrimental to scattering performance, likely due to overfitting in the high-dimensional feature space.
Conversely, enforcing rotation invariance proves beneficial.
Regarding metric reliability, training marginal likelihood is misleading when ARD is included, as it favors the overfitting ARD models (low train loss, high test loss).
However, if ARD configurations are excluded, the training loss correctly identifies that neural network features are superior to scattering features, aligning with test performance.
Finally, consistent with other tasks, deep ensembles provide reasonable mean predictions but fail to deliver reliable uncertainty estimates.

\subsubsection{More Data and Approximate Inference}

\begin{table*}[t]
\centering
\caption{Performance comparison on the full datasets using SVGP inference. We report mean and standard deviation across 5 training runs on the fixed full split. Best results are \underline{underlined}; those not statistically distinguishable are \textbf{bolded}.}
\label{tab:more_data}
\small
\setlength{\tabcolsep}{0pt} 
\begin{tabular*}{\linewidth}{@{\extracolsep{\fill}} llcccccc}
\toprule
\multirow{2}{*}[-0.5ex]{\textbf{Data}} & \multirow{2}{*}[-0.5ex]{\textbf{Metric}} & \multirow{2}{*}[-0.5ex]{\textbf{Trivial}} & \multirow{2}{*}[-0.5ex]{\textbf{Scattering}} & \multicolumn{2}{c}{\textbf{ConvNeXt}} & \multicolumn{2}{c}{\textbf{DINOv2}} \\
\cmidrule(lr){5-6} \cmidrule(lr){7-8}
 & & & & \textbf{Atto} & \textbf{Base} & \textbf{Small} & \textbf{Base} \\
\midrule

\multirow{5}{*}{\textbf{SLP}}
& RMSE            & 817 $\pm$ 35 & $\mathbf{540 \pm 33}$ & \underline{$\mathbf{515 \pm 39}$} & $\mathbf{556 \pm 21}$ & 748 $\pm$ 56 & 687 $\pm$ 40 \\
& NLL             & 1.50 $\pm$ 0.05 & $\mathbf{1.16 \pm 0.12}$ & \underline{$\mathbf{1.13 \pm 0.13}$} & $\mathbf{1.15 \pm 0.07}$ & 1.46 $\pm$ 0.12 & 1.40 $\pm$ 0.10 \\
& QCE             & 0.03 $\pm$ 0.01 & 0.08 $\pm$ 0.02 & 0.08 $\pm$ 0.02 & \underline{\textbf{0.06 $\pm$ 0.02}} & 0.10 $\pm$ 0.02 & 0.10 $\pm$ 0.02 \\
& PI-$\mu$        & 3.92 $\pm$ 0.00 & \underline{$\mathbf{1.84 \pm 0.01}$} & 1.93 $\pm$ 0.01 & 2.32 $\pm$ 0.05 & 2.71 $\pm$ 0.08 & 2.50 $\pm$ 0.09 \\
& PI-$\sigma$     & 0.00 $\pm$ 0.00 & 0.11 $\pm$ 0.00 & 0.09 $\pm$ 0.01 & 0.07 $\pm$ 0.01 & 0.13 $\pm$ 0.01 & 0.17 $\pm$ 0.04 \\

\midrule

\multirow{5}{*}{\textbf{HP}}
& RMSE            & 623 $\pm$ 14 & \underline{$\mathbf{375 \pm 12}$} & 451 $\pm$ 25 & 481 $\pm$ 26 & 421 $\pm$ 19 & $\mathbf{392 \pm 20}$ \\
& NLL             & 1.31 $\pm$ 0.02 & $\mathbf{0.82 \pm 0.02}$ & 0.87 $\pm$ 0.03 & 0.92 $\pm$ 0.03 & $\mathbf{0.84 \pm 0.03}$ & \underline{$\mathbf{0.77 \pm 0.04}$} \\
& QCE             & 0.05 $\pm$ 0.00 & 0.02 $\pm$ 0.01 & 0.03 $\pm$ 0.01 & 0.03 $\pm$ 0.01 & \underline{\textbf{0.00 $\pm$ 0.01}} & \underline{\textbf{0.00 $\pm$ 0.01}} \\
& PI-$\mu$        & 3.92 $\pm$ 0.00 & 2.56 $\pm$ 0.00 & 2.45 $\pm$ 0.03 & 2.58 $\pm$ 0.03 & 2.55 $\pm$ 0.03 & \underline{$\mathbf{2.36 \pm 0.03}$} \\
& PI-$\sigma$     & 0.00 $\pm$ 0.00 & 0.02 $\pm$ 0.00 & 0.05 $\pm$ 0.01 & 0.06 $\pm$ 0.00 & 0.11 $\pm$ 0.03 & 0.12 $\pm$ 0.01 \\

\midrule

\multirow{5}{*}{\textbf{AW}}
& RMSE            & 0.90 $\pm$ 0.02 & 0.56 $\pm$ 0.04 & $\mathbf{0.50 \pm 0.01}$ & $\mathbf{0.50 \pm 0.03}$ & $\mathbf{0.49 \pm 0.01}$ & \underline{$\mathbf{0.49 \pm 0.01}$} \\
& NLL             & 1.54 $\pm$ 0.03 & 0.96 $\pm$ 0.05 & \underline{$\mathbf{0.86 \pm 0.02}$} & $\mathbf{0.88 \pm 0.05}$ & $\mathbf{0.87 \pm 0.03}$ & $\mathbf{0.87 \pm 0.03}$ \\
& QCE             & 0.01 $\pm$ 0.01 & 0.01 $\pm$ 0.01 & \underline{\textbf{0.01 $\pm$ 0.01}} & \textbf{0.01 $\pm$ 0.01} & \textbf{0.01 $\pm$ 0.01} & \textbf{0.01 $\pm$ 0.01} \\
& PI-$\mu$        & 3.92 $\pm$ 0.00 & 2.79 $\pm$ 0.06 & $\mathbf{2.30 \pm 0.01}$ & 2.42 $\pm$ 0.02 & 2.34 $\pm$ 0.01 & \underline{$\mathbf{2.28 \pm 0.01}$} \\
& PI-$\sigma$     & 0.00 $\pm$ 0.00 & 0.10 $\pm$ 0.03 & 0.08 $\pm$ 0.00 & 0.03 $\pm$ 0.00 & 0.04 $\pm$ 0.00 & 0.04 $\pm$ 0.00 \\

\bottomrule
\end{tabular*}
\end{table*}

To assess scalability and performance on larger datasets, we employ SVGP inference.
We conduct this evaluation on the three image regression datasets considered.
The results are summarized in \Cref{tab:more_data}.
Detailed tabulated results are provided in the file \texttt{dist-shift-svgp.pdf} in \url{https://github.com/nash169/bayesian-scattering/tree/main/results}.
Note that unlike the experiments in the main text, the the train test here is fixed; thus, the reported variance across 5 seeds arises from the stochastic optimization, initialization, and randomness in test set subsampling (we chose 250 random test points out of the whole test set for every trial).

\subsection{Bayesian Optimization}

\begin{figure}[h]
	\centering
	\subfloat[][\texttt{QM2D}, Matérn-$\nicefrac{5}{2}$]{\includegraphics[height=3.9cm]{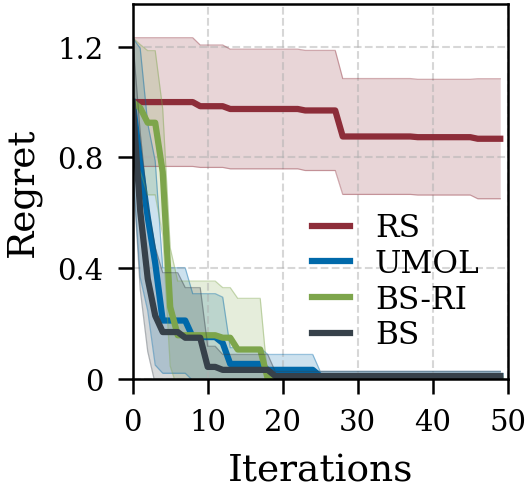}\label{fig:bo_comparison_extended:2d:matern}}
    \hfill
    \subfloat[][\texttt{QM2D}, Matérn-$\nicefrac{5}{2}$-ARD]{\includegraphics[height=3.9cm]{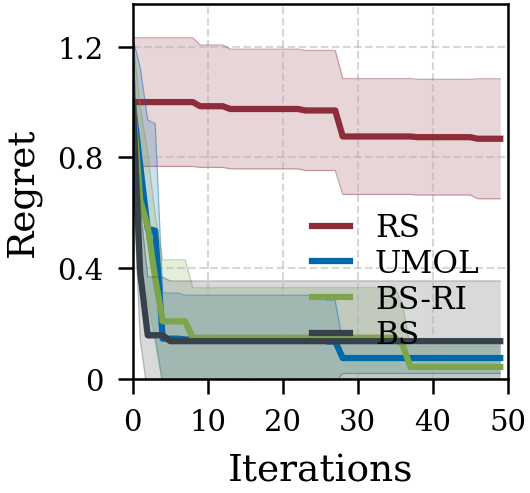}\label{fig:bo_comparison_extended:2d:matern_ard}}
    \hfill
	\subfloat[][\texttt{QM2D}, RBF]{\includegraphics[height=3.9cm]{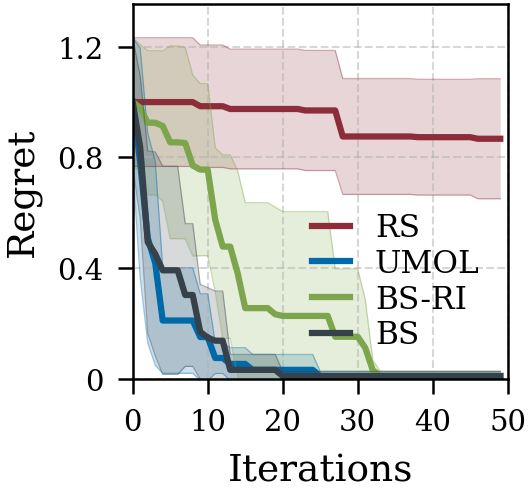}\label{fig:bo_comparison_extended:2d:rbf}}
    \hfill
	\subfloat[][\texttt{QM2D}, RBF-ARD]{\includegraphics[height=3.9cm]{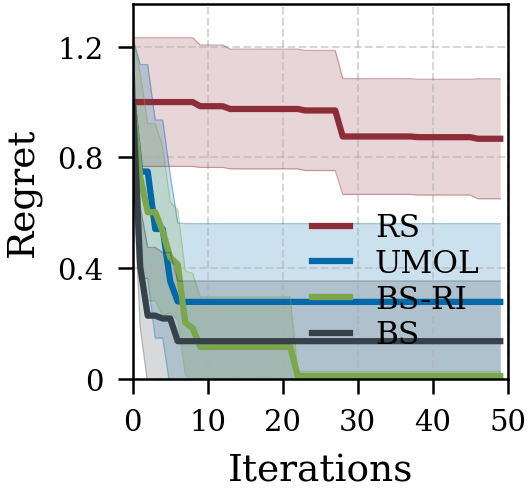}\label{fig:bo_comparison_extended:2d:rbf_ard}}
    \\[\baselineskip]
	\subfloat[][IP on \texttt{QM9}, Matérn-$\nicefrac{5}{2}$]{\includegraphics[height=3.9cm]{figures/bo/qm9_alpha_gp_exact_reg_matern52_bo_95ci}\label{fig:bo_comparison_extended:3d:ip:matern}}
    \hfill
    \subfloat[][IP on \texttt{QM9}, Matérn-$\nicefrac{5}{2}$-ARD]{\includegraphics[height=3.9cm]{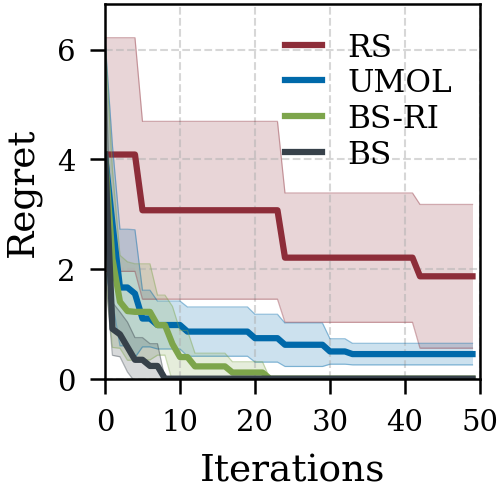}\label{fig:bo_comparison_extended:3d:ip:matern_ard}}
    \hfill
	\subfloat[][IP on \texttt{QM9}, RBF]{\includegraphics[height=3.9cm]{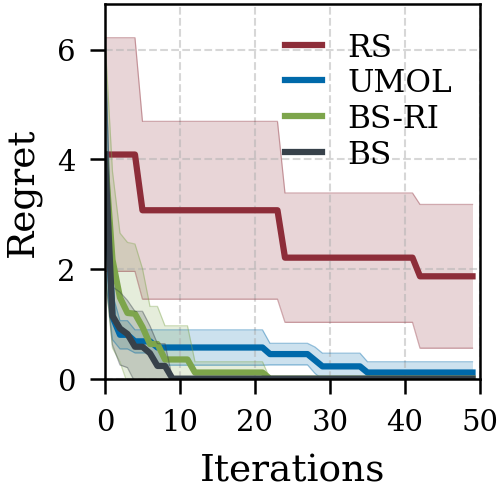}\label{fig:bo_comparison_extended:3d:ip:rbf}}
    \hfill
	\subfloat[][IP on \texttt{QM9}, RBF-ARD]{\includegraphics[height=3.9cm]{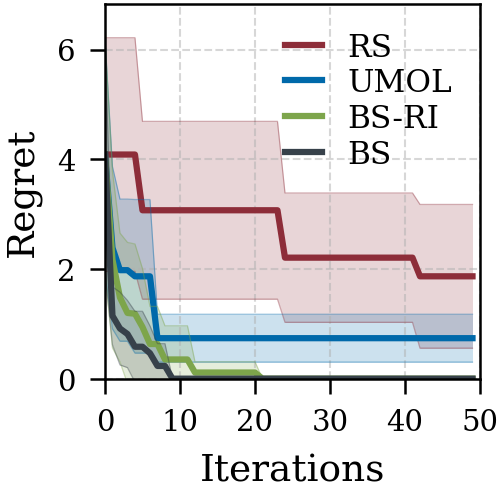}\label{fig:bo_comparison_extended:3d:ip:rbf_ard}}
    \\[\baselineskip]
	\subfloat[][ESE on \texttt{QM9}, Matérn-$\nicefrac{5}{2}$]{\includegraphics[height=3.9cm]{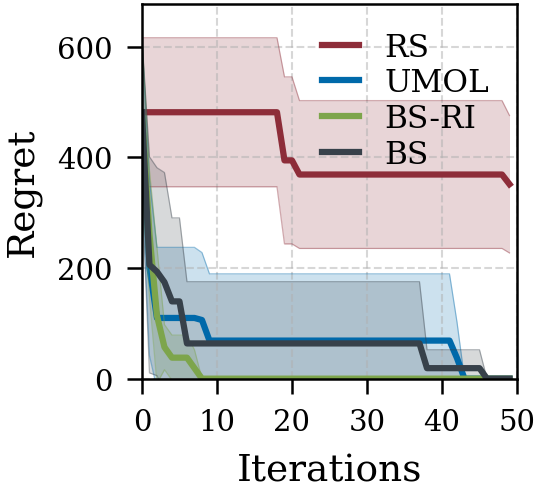}\label{fig:bo_comparison_extended:3d:ese:matern}}
    \hfill
    \subfloat[][ESE on \texttt{QM9}, Matérn-$\nicefrac{5}{2}$-ARD]{\includegraphics[height=3.9cm]{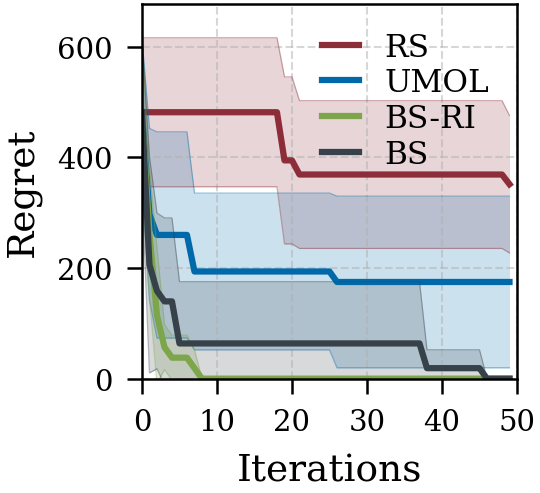}\label{fig:bo_comparison_extended:3d:ese:matern_ard}}
    \hfill
	\subfloat[][ESE on \texttt{QM9}, RBF]{\includegraphics[height=3.9cm]{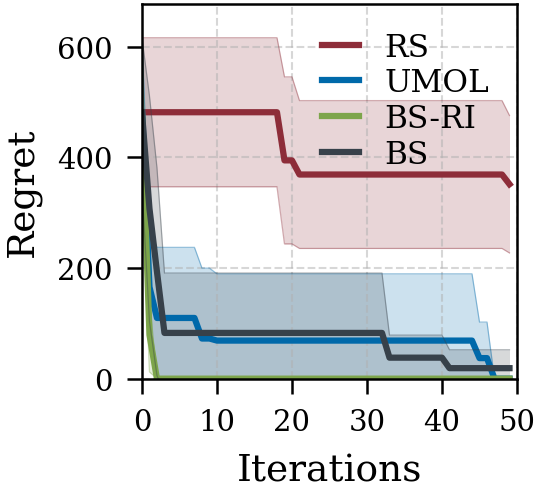}\label{fig:bo_comparison_extended:3d:ese:rbf}}
    \hfill
	\subfloat[][ESE on \texttt{QM9}, RBF-ARD]{\includegraphics[height=3.9cm]{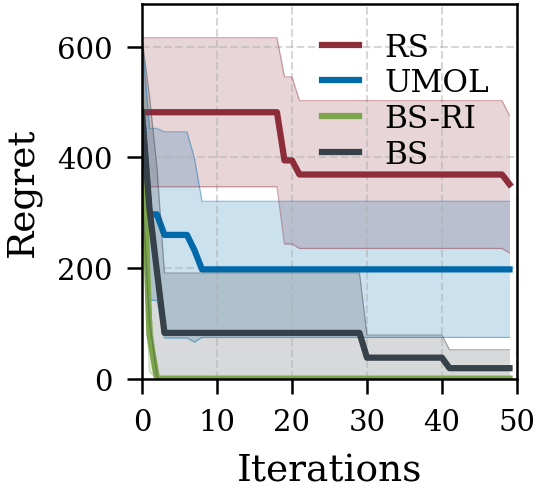}\label{fig:bo_comparison_extended:3d:ese:rbf_ard}}
    \\[\baselineskip]
	\subfloat[][U0 on \texttt{QM9}, Matérn-$\nicefrac{5}{2}$]{\includegraphics[height=3.9cm]{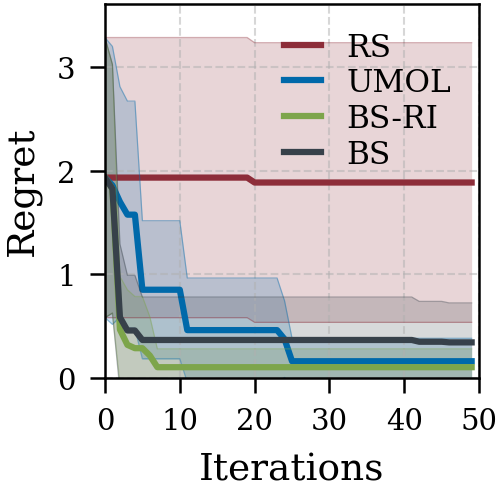}\label{fig:bo_comparison_extended:3d:u0:matern}}
    \hfill
    \subfloat[][U0 on \texttt{QM9}, Matérn-$\nicefrac{5}{2}$-ARD$\!\!\!\!$]{\includegraphics[height=3.9cm]{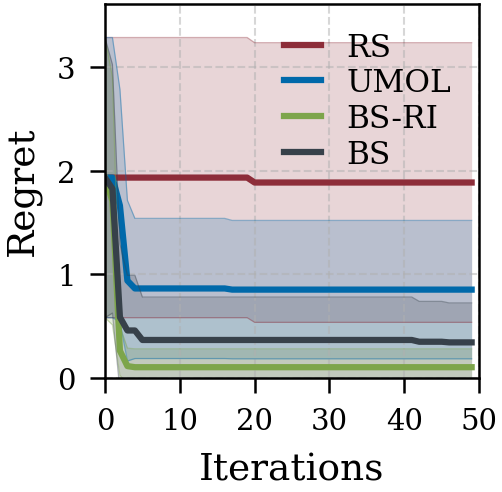}\label{fig:bo_comparison_extended:3d:u0:matern_ard}}
    \hfill
	\subfloat[][U0 on \texttt{QM9}, RBF]{\includegraphics[height=3.9cm]{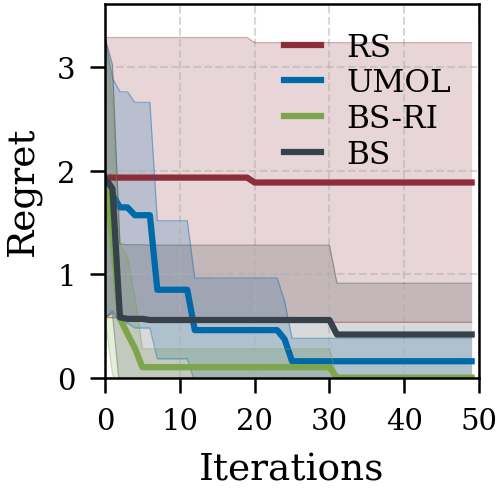}\label{fig:bo_comparison_extended:3d:u0:rbf}}
    \hfill
	\subfloat[][U0 on \texttt{QM9}, RBF-ARD]{\includegraphics[height=3.9cm]{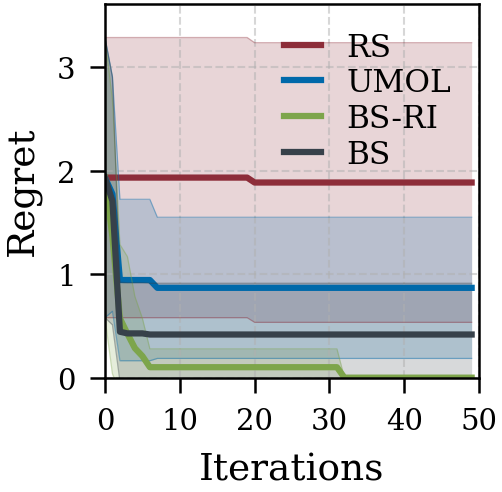}\label{fig:bo_comparison_extended:3d:u0:rbf_ard}}
    \caption{
    Molecular property BO (extended).
    Curves show the mean regret over 5 runs; shaded regions show $0.95$ confidence intervals.
    }
    \label{fig:bo_comparison_extended}
\end{figure}

Regret plots for Bayesian optimization with different probabilistic heads can be found in~\Cref{fig:bo_comparison_extended}.

\end{document}

%% file: figures/intro/figure.tex
\definecolor{techblue}{RGB}{0, 51, 102} 
\definecolor{techaccent}{RGB}{70, 130, 180} 

\begin{tikzpicture}[
    node distance=1.2cm,
    font=\small\sffamily,
    >=Stealth,
    arrow/.style={
        ->, 
        thick, 
        color=techblue,
        shorten <=3pt, 
        shorten >=3pt
    },
    label_text/.style={
        align=center, 
        font=\small, 
        anchor=north
    },
    arrow label top/.style={
        midway, above, align=center, font=\small\bfseries, text=black, yshift=2pt
    },
    arrow label bottom/.style={
        midway, below, align=center, font=\footnotesize, color=gray!80!black, yshift=-2pt
    }
]

    \node (input) [inner sep=0pt, minimum size=1.8cm] {
        \includegraphics[width=1.8cm, keepaspectratio]{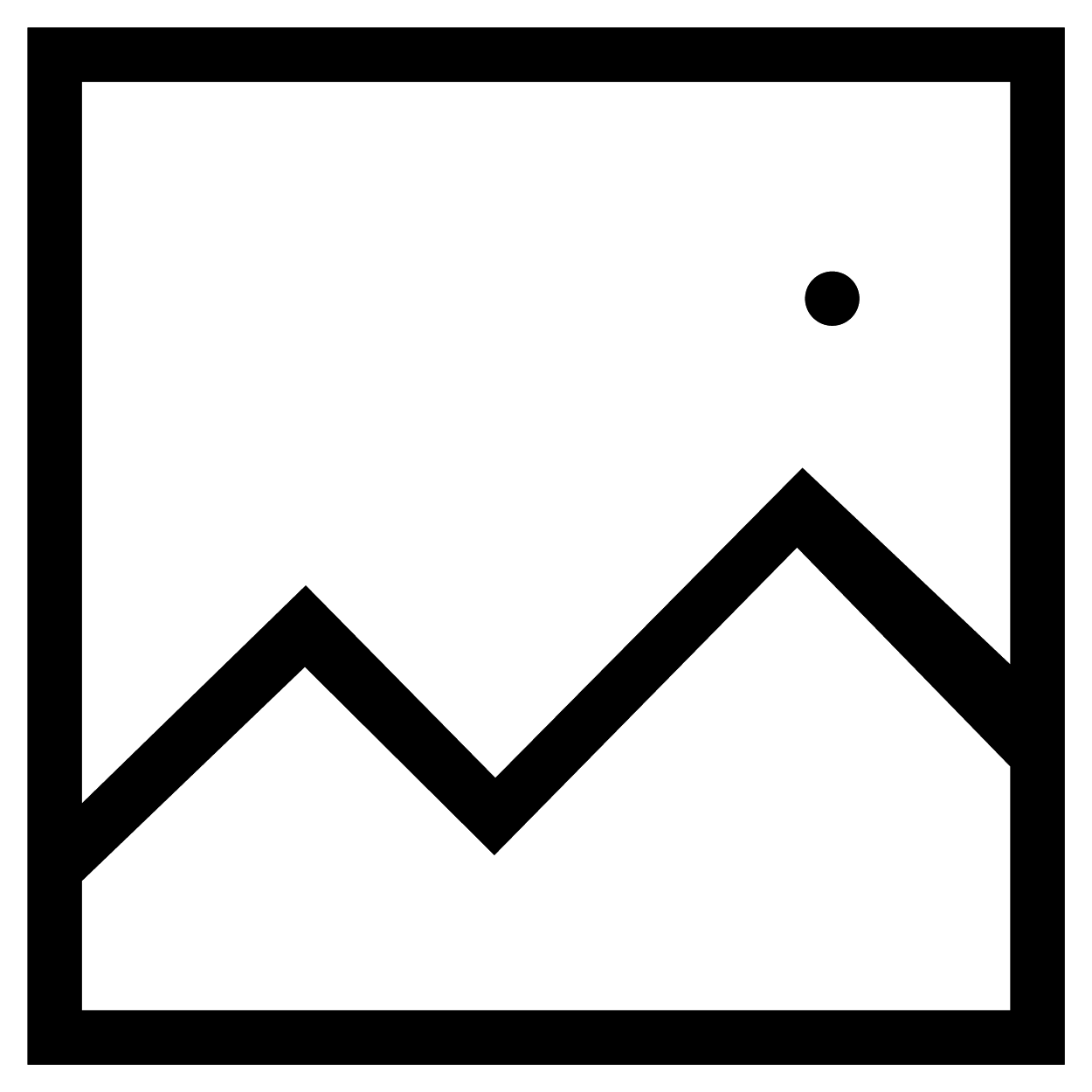}
    };

    \node (features) [right=5cm of input] {
        \begin{tikzpicture}[scale=0.45, baseline=(current bounding box.center)]
            \foreach \intensity [count=\y] in {10, 60, 25, 90, 50, 15, 80, 35} {
                \fill[techblue!\intensity] (0, \y*0.5) rectangle (1, \y*0.5+0.5);
                \draw[white, thin] (0, \y*0.5) -- (1, \y*0.5); 
            }
            \draw[techblue, thin] (0, 0.5) rectangle (1, 4.5);
        \end{tikzpicture}
    };

    \draw[arrow] (input) -- 
        node[arrow label top] {Scattering Transform} 
        node[arrow label bottom] {
            Wavelet transform $\psi$ \\ 
            + Modulus $|\cdot|$ \\ 
            + Averaging $\int$
        } (features);

    \node (output) [right=4.5cm of features] {
        \begin{tikzpicture}[
            scale=0.9, 
            baseline=(current bounding box.center),
            declare function={gauss(\x)=1.8*exp(-(\x-1.5)^2/0.2);}
        ]
            \draw[->, gray, thin] (0,0) -- (3.0,0) node[right, font=\footnotesize] {$y$};
            \draw[->, gray, thin] (0,0) -- (0,2.2) node[left, font=\footnotesize] {$P$};
            
            \fill[techaccent, opacity=0.3] plot[domain=0.0:2.8, samples=50] (\x, {gauss(\x)}) -- cycle;
            \draw[techblue, thick]         plot[domain=0.0:2.8, samples=50] (\x, {gauss(\x)});
            \draw[dashed, techblue, thin] (1.5,0) -- (1.5, 1.8);
        \end{tikzpicture}
    };

    \draw[arrow] (features) -- 
        node[arrow label top] {Probabilistic Head} 
        node[arrow label bottom] {
            Gaussian process model \\
            + Standard kernel \\
            ~~~~(RBF, Matérn)
        } (output);

    \coordinate (label_baseline) at (0, -1.6); 
    \node[label_text] at (input |- label_baseline) {Input image $\mathbf{X}$};
    \node[label_text] at (features |- label_baseline) {Features $\Phi(\mathbf{X})$};
    \node[label_text] at (output |- label_baseline) {Probabilistic prediction $p(y|\Phi(\mathbf{X}))$};

\end{tikzpicture}